\documentclass[journal,twoside,web]{ieeecolor}
\usepackage{tmi}
\usepackage{url}
\usepackage{cite}
\usepackage{soul}
\usepackage{float}
\usepackage{array}
\usepackage{dsfont}
\usepackage{xcolor}
\usepackage{comment}
\usepackage{multirow}
\usepackage{graphicx}
\usepackage{mathrsfs}
\usepackage{textcomp}
 
\usepackage{booktabs}
\usepackage{algorithmic}
\usepackage{amsmath,amssymb,amsfonts}

\def\ie{\textit{i.e.}}
\def\eg{\textit{e.g.}}
\def\etal{{\textit{et~al.}}}

\newcommand{\myparagraph}[1]{\vspace{3pt}\noindent{\bf{#1}}~~}
\newcommand\ti[1]{\textit{#1}}
\newcommand\mf[1]{\mathbf{#1}}

\newcommand{\ours}{SimCVD}

\let\labelindent\relax
\usepackage{enumitem}

\def\BibTeX{{\rm B\kern-.05em{\sc i\kern-.025em b}\kern-.08em
    T\kern-.1667em\lower.7ex\hbox{E}\kern-.125emX}}
\begin{document}
\title{\ours: Simple Contrastive Voxel-Wise Representation Distillation for Semi-Supervised Medical Image Segmentation}
\author{Chenyu You, Yuan Zhou, Ruihan Zhao, Lawrence Staib, James S. Duncan, \IEEEmembership{Life Fellow, IEEE}
\thanks{C. You is with the Department of Electrical Engineering, Yale University, New Haven, CT 06510 USA (e-mail:chenyu.you@yale.edu).}
\thanks{Y. Zhou is with the Department of Radiology \& Biomedical Imaging, Yale University, New Haven, CT 06510 USA.}
\thanks{R. Zhao is with the Department of Electrical and Computer Engineering, The University of Texas at Austin, TX 78712 USA.}
\thanks{L. H. Staib and J. S. Duncan are with Departments of Radiology \& Biomedical Imaging, Biomedical Engineering, and Electrical Engineering, Yale University, New Haven, CT 06511 USA (e-mail:james.duncan@yale.edu).}
}

\maketitle

\begin{abstract}

Automated segmentation in medical image analysis is a challenging task that requires a large amount of manually labeled data. However, most existing learning-based approaches usually suffer from limited manually annotated medical data, which poses a major practical problem for accurate and robust medical image segmentation. In addition, most existing semi-supervised approaches are usually not robust compared with the supervised counterparts, and also lack explicit modeling of geometric structure and semantic information, both of which limit the segmentation accuracy. In this work, we present {\ours}, a simple contrastive distillation framework that significantly advances state-of-the-art voxel-wise representation learning. We first describe an unsupervised training strategy, which takes two views of an input volume and predicts their signed distance maps of object boundaries in a contrastive objective, with only two independent dropout as mask. This simple approach works surprisingly well, performing on the same level as previous fully supervised methods with much less labeled data. We hypothesize that dropout can be viewed as a minimal form of data augmentation and makes the network robust to representation collapse. Then, we propose to perform structural distillation by distilling pair-wise similarities. We evaluate {\ours} on two popular datasets: the Left Atrial Segmentation Challenge (LA) and the NIH pancreas CT dataset. The results on the LA dataset demonstrate that, in two types of labeled ratios (\ie, 20\% and 10\%), {\ours} achieves an average Dice score of 90.85\% and 89.03\% respectively, a 0.91\% and 2.22\% improvement compared to previous best results. Our method can be trained in an end-to-end fashion, showing the promise of utilizing {\ours} as a general framework for downstream tasks, such as medical image synthesis, enhancement, and registration.


\end{abstract}
\section{Introduction}
\label{sec:intro}


\begin{figure*}[!t]
    \centering
    \includegraphics[width=0.8\textwidth]{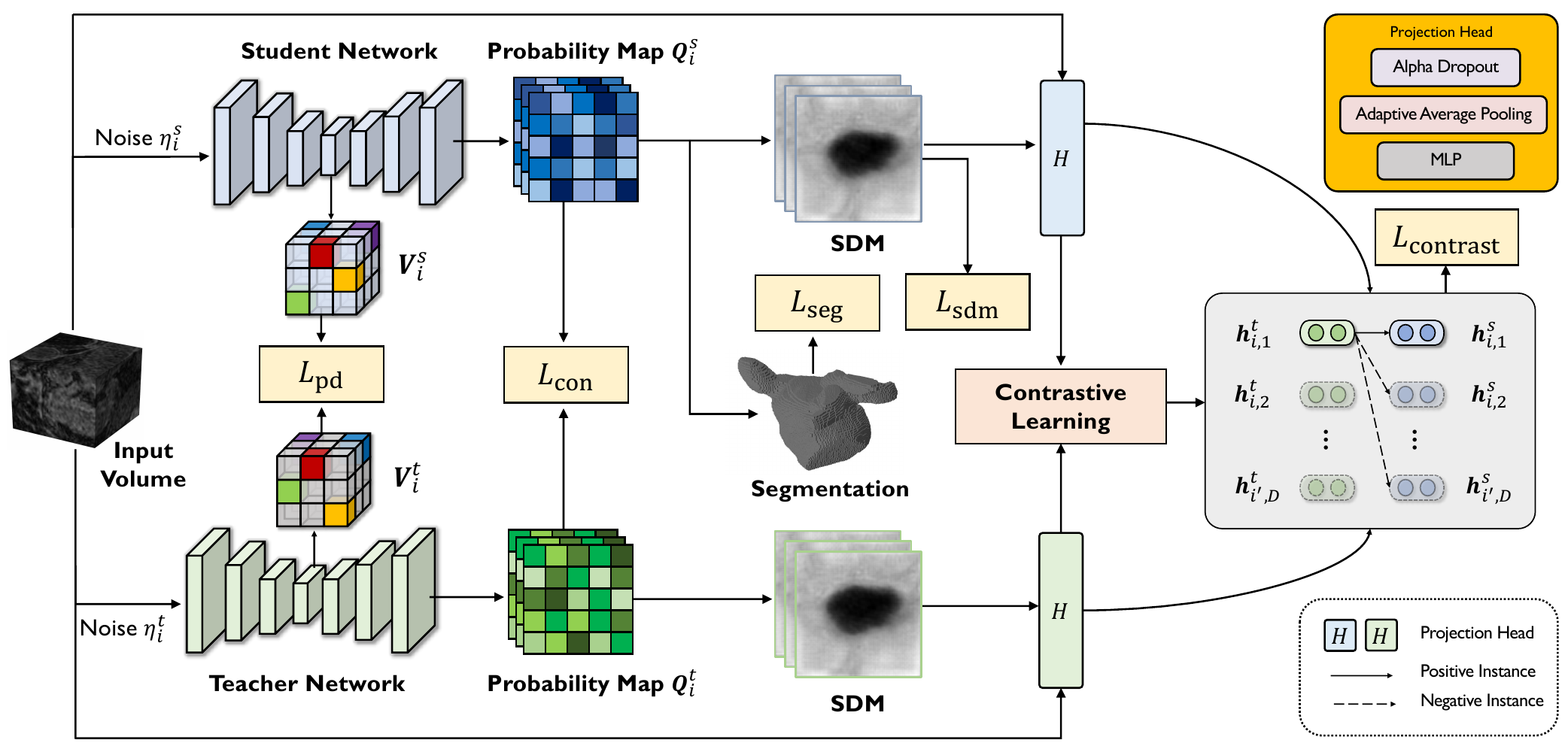}
    \vspace{-5pt}
    \caption{Overview of our {\ours} in training. Given a 3D input volume, our SimCVD jointly predicts the 3D probability maps and the SDMs of the object using a student network and a teacher network. The student network is trained by stochastic gradient descent using two supervised losses ($\mathcal{L}_{\mathrm{seg}}$, $\mathcal{L}_{\mathrm{sdm}}$) and three unsupervised terms ($\mathcal{L}_{\mathrm{contrast}}$, $\mathcal{L}_{\mathrm{pd}}$, 
    $\mathcal{L}_{\mathrm{con}}$). Specifically, $\mathcal{L}_{\mathrm{contrast}}$ is designed to distill ``boundary-aware'' knowledge by contrastive learning in the shared latent space, and $\mathcal{L}_{\mathrm{pd}}$ is designed to exploit structural relationships among location-paired voxel representations from the encoders. The teacher network’s weights are updated with a ``momentum update'' (exponential moving average) of the student network’s weights.}
    \label{fig:structure}
    \vspace{-15pt}
\end{figure*}

Medical image segmentation is a popular task in both machine learning and medical imaging communities \cite{staib1996model,yang2004neighbor,yang20043d,chakraborty1996deformable,staib1992boundary}. Compared to traditional segmentation approaches, deep neural network based segmentation methods have achieved much stronger performance in recent years with huge advances in representation learning \cite{ronneberger2015u,milletari2016v,bai2017semi,ganaye2018semi,you2020unsupervised,wang2020deep,xue2020shape,li2020transformation}. However, previous state-of-the-art approaches are mostly trained with a large amount of labeled data, which pose significant practical challenges in many medical segmentation tasks where there is a scarcity of labeled data due to  the heavy burden of annotating images. 

In recent years, a wide variety of semi-supervised methods \cite{laine2016temporal,zhang2017deep,li2018semi,nie2018asdnet,qiao2018deep,bortsova2019semi,li2021assessing,yu2019uncertainty,li2020shape,zhu2020rubik,luo2020semi,chaitanya2020contrastive,you2021momentum} have been designed to tackle these issues, which learn from limited labeled data along with a large amount of unlabeled data, achieving significant improvements in accuracy and greatly reducing the labeling cost. The common paradigms include adversarial learning, knowledge distillation, and self-supervised learning. Contrastive learning, a sub-area of self-supervised learning, has recently been noted as a promising direction since it has shown great promise in learning useful representations with limited human supervision~\cite{chen2020simple,hjelm2018learning,bai2019self,chaitanya2020contrastive,peng2021self}. This is often best understood as pulling together semantically similar (\ti{positive}) samples and pushing apart non-similar (\ti{negative}) samples in a shared latent space. The representations uncovered by these contrastive objectives are capable of boosting the performance of any vision system especially in scenarios where the amount of annotated data available for the downstream tasks is extremely low, which is well suited for medical image analysis.

Despite advances in semi-supervised learning benchmarks, previous methods still face several major challenges: (1) \ti{Suboptimal performance}: although prior works have achieved promising segmentation accuracy in the setting of limited annotations, semi-supervised models are usually not robust due to some information loss, compared with fully-supervised counterparts; (2) \ti{Geometric information loss}: previous segmentation networks are poor at characterizing geometry, \ie, leveraging the intrinsic geometric structure of the images, such as the object boundary. As a consequence, it is often hard to accurately recognize object contours; and (3) \ti{Generalization ability}: considering the limited amount of training data, training deep models is usually deficient due to over-fitting and co-adapting \cite{yang2019snapshot,zhuang2020deep}.

In this work, we address the question: can we advance state-of-the-art voxel-wise representation learning in a more extreme few-annotation-setting for medical image segmentation? To this end, we present {\ours}, a \underline{sim}ple \underline{c}ontrastive \underline{v}oxel-wise representation \underline{d}istillation framework, which can be utilized to produce superior voxel-wise representations from unlabeled data for improving network performance. Our proposed {\ours}, built upon the mean-teacher framework \cite{tarvainen2017mean}, can address the above-mentioned challenges as follows. First, {\ours} predicts the output geometric representations with only two different \ti{dropout}~\cite{srivastava2014dropout} masks (Figure~\ref{fig:structure}). In other words, we pass two views of the geometric representations to the mean-teacher model, obtain two representations as ``positive pairs'', by applying two independent dropout masks, and learn effective representations by efficiently associating positives and disassociating negatives in the shared latent space. Though this unsupervised learning strategy is simple, we find this approach is strikingly effective compared to other common data augmentation techniques (\eg, inpainting and local shuffle pixel). More importantly, as we will show, it achieves comparable performance to previous fully-supervised approaches. Through a series of thorough analyses, we find that dropout can be viewed as minimal data augmentation for performance improvement, and it can effectively regularize the training of deep neural networks, avoid representation collapse and enhance model generalization.

Second, we attribute the cause of the \ti{geometric information loss} to the need for geometric shape constraints. We address this challenge by performing multi-task learning that jointly predicts a segmentation map along with a signed distance map (SDM) \cite{perera2015motion,dangi2019distance, park2019deepsdf,xue2020shape,li2020shape}. The SDM calculates the signed distance function of the object, \ie, the distance of a voxel from the boundary of the object, with the sign determined by whether the voxel is within the object. Thus, it can be viewed as a global shape constraint on the labeled data. Considering that the SDM can provide a more flexible geometric measure of the object boundary, we move beyond the supervised learning scheme and exploit the regularity in geometric shapes among different object classes through distilling ``boundary-aware'' knowledge via a contrastive objective among the unlabeled data. This enables the model to learn boundary-aware features more effectively by encouraging the networks to produce segmentation maps with similar distance map distributions on the entire dataset.

Third, it is challenging to train the segmentation model on small training sets since deep neural networks trained on a limited amount of data are prone to over-fitting. To this end, we propose to use knowledge distillation (KD), which has been shown to be effective in segmentation and classification tasks \cite{hinton2015distilling,romero2014fitnets,liu2019structured}. The key idea of KD is that a teacher model is first trained, and then used to guide the training of the student model for improving generalization ability. In the medical domain, most existing KD methods~\cite{yu2019annotation,li2020dual} simply consider the segmentation problem as a pixel/voxel-level classification problem. In contrast, considering that medical image semantic segmentation is a structured prediction problem, we present a novel structured knowledge \ti{pair-wise distillation}, which further use the structural knowledge from the mean-teacher model, while avoiding co-adapting and over-fitting.

Our contributions are summarized as follows. First, we propose a novel contrastive distillation model termed {\ours} featured by (i) boundary-aware representations that incorporate rich information of the object shape, (ii) a distillation objective which contrasts different distance map distributions jointly in the shared latent space, and (iii) a pair-wise distillation objective to further distill pair-wise structural knowledge. Second, we demonstrate that, in the setting of very limited annotation, simply using dropout can deliver more robust end-to-end segmentation performance compared to heavily relying on a large amount of labeled data. Third, we conduct experiments on two popular benchmark datasets to evaluate {\ours}. The results demonstrate that {\ours} significantly outperforms other state-of-the-art semi-supervised approaches, while achieving competitive performance compared to fully-supervised counterparts.

\section{Related Work}

\myparagraph{Semi-Supervised Medical Image Segmentation}
In recent years, substantial efforts~\cite{zhang2017deep,li2018semi,nie2018asdnet,he2019dpa,zhou2019models,yang2020nuset,laine2016temporal,yu2019uncertainty,chen2021deep,shanlin2022CVPR,tang2019clinically,tang2021spatial,zhang2018fully,zhang2021fully,tang2019nodulenet,zhang2018fully,zhang2020fully,sun2020attentionanatomy,yan2022after,tang2021recurrent,xue2020shape,bortsova2019semi,ma2020distance,castillo2020auxiliary,murugesan2019psi,xia2020uncertainty,zheng2019semi,zhuang2019self,taleb20203d,you2022class} have been devoted to incorporating unlabeled data to improve network performance due to limited annotations. Yu~\etal\cite{yu2019uncertainty} investigated an uncertainty map based on the mean-teacher framework~\cite{tarvainen2017mean} to guide the student network to capture better features. Li~\etal\cite{li2020shape} proposed to use signed distance fields for boundary prediction to improve the performance. Also, Luo~\etal\cite{luo2020semi} proposed a dual-task-consistency (DTC) model for semi-supervised medical image segmentation by jointly predicting the pixel-wise segmentation maps and the global-level level set representations on the unlabeled data. Our method aims at a more practical and challenging scenario: we train our model in a more extreme few-annotation setting that relies only on a small number of annotations, while achieving superior segmentation accuracy.

\myparagraph{Contrastive Learning}
Self-supervised learning (SSL)~\cite{hadsell2006dimensionality,doersch2015unsupervised,noroozi2016unsupervised,zhuang2019self} has provided robust benefits to vision tasks by learning effective visual representations from unlabeled data in an unsupervised setting. It is based on a commonly-held belief that superior performance gains can be achieved through improved representation learning. Recently, contrastive learning, a type of self-supervised learning, has received a lot of interests~\cite{hadsell2006dimensionality,wu2018unsupervised,tian2019contrastive,misra2020self,chen2020simple,Federici2020Learning,chaitanya2020contrastive,peng2021self,chen2021self,you2021self}. The key idea of contrastive learning is to learn powerful representations that optimize similarity constraints to discriminate similar pairs (\ti{positive}) and dissimilar pairs (\ti{negative}) within a dataset. The primary stream of subsequent work focuses on the choice of dissimilar pairs, which is critical to the quality of learned representations. The loss function used to quantify the contrast is chosen from several options, such as InfoNCE~\cite{oord2018representation}, Triplet~\cite{wang2015unsupervised}, and so on. Recent studies~\cite{wu2018unsupervised,misra2020self} introduced memory bank or momentum contrast to use more negative samples for contrast computation. In the context of medical imaging, Chaitanya~\etal\cite{chaitanya2020contrastive} extended a contrastive learning framework to extract global and local cues in a stage-wise way, which requires human intervention and extensive training time. In contrast to Chaitanya~\etal\cite{chaitanya2020contrastive}, our unified work focuses on explicit modeling of the intrinsic geometric structure of the semantic objects in an end-to-end manner, and hence is able to recognize object boundaries more effectively and efficiently.

\myparagraph{Knowledge Distillation}
The idea of knowledge distillation is to minimize the KL-divergence between the output distributions of the teacher model and the student model, and thus avoid over-fitting. KD has been applied to a variety of tasks \cite{you2020data,you2020contextualized,you2021mrd,you2021knowledge,ma2020undistillable,ma2021good}, including image classification \cite{hinton2015distilling,furlanello2018born,yang2018knowledge,li2017learning} and semantic segmentation \cite{xie2018improving,liu2019structured}. Recent works \cite{yang2018knowledge,furlanello2018born} found that the student model can outperform the teacher model when they share the same network architecture. Zhang~\etal\cite{zhang2018deep} proposed to collaboratively train multiple student models with co-distillation, which improves performance of those individual models. At the same time, in the context of medical imaging, among those existing state-of-the-art KD methods, the self-ensemble mean-teacher framework \cite{tarvainen2017mean} is widely explored for image segmentation. Different from the existing methods that separately exploit class probabilities for each voxel, we consider knowledge distillation as a structured prediction problem by matching the relational similarity among all pairs of voxels from the encoded feature maps of the mean-teacher model. We have found that our approach significantly improves learning better voxel-wise representations.
\section{Method}
\label{sec:method}

In this section, we introduce {\ours}, a semi-supervised segmentation network, which is built from scratch by effectively leveraging scarce labeled data and ample unlabeled data for improving end-to-end voxel-wise representation learning (See Figure~\ref{fig:structure}). We first overview our proposed {\ours} and then describe the task formulation of {\ours}. Finally, we detail each component of {\ours} in the following subsections. 

\subsection{Overview}

We aim to construct an end-to-end voxel-wise contrastive distillation algorithm to learn boundary-aware representations in the setting of extremely few annotations for volumetric medical imaging segmentation. Although the accuracy of supervised models is usually higher than that of semi-supervised models, the former requires much more labeled data than the latter. In many clinical situations, we only have few annotated data but a large amount of unlabeled data. This situation necessitates a semi-supervised segmentation algorithm that can utilize the unlabeled data to improve the segmentation performance.

To this end, we propose a novel contrastive distillation framework to advance state-of-the-art voxel-wise representation learning. In particular, our base multi-task segmentation network tackles two tasks simultaneously: classification and regression. Specifically, the segmentation network takes the input volume batch and jointly predicts the probability maps (classification) and the SDMs of the object (regression). To obtain better representations, we propose to perform structured distillation in the latent \ti{feature} space, followed by contrasting the boundary-aware features in the \ti{prediction} space, to learn more effective boundary-aware representations from 3D unlabeled data by regularizing the embedding space and exploring the geometric and spatial context of training voxels. At test time, we remove the mean teacher and two projection heads, and only the student network is deployed for the medical segmentation tasks. 

\subsection{Task Formulation}
In this work, we consider a set of training data (3D images) including $N$ labeled data and $M$ unlabeled data, where $N\ll M$. For simplicity, we denote the small set of labeled data as $\mathcal{D}_{l}=\left\{\left(\mathbf{X}_{i}, \mathbf{Y}_{i},\mathbf{Y}^{\mathrm{sdm}}_{i}\right)\right\}_{i=1}^{N}$, and abundant unlabeled data as $\mathcal{D}_{u}=\{\mathbf{X}_{i}\}_{i=N+1}^{N+M}$, where~$\mathbf{X}_{i} \in \mathbb{R}^{H \times W \times D}$ is the volume input, $\mathbf{Y}_{i} \in \{0, 1\}^{H \times W \times D}$ is the ground-truth label, and $\mathbf{Y}^{\mathrm{sdm}}_i \in \mathbb{R} ^{H \times W \times D}$ is the computed ground truth SDMs from $\mathbf{Y}_i$, which measures the distance from each voxel to the object boundary. Every 3D image $\mathbf{X}_{i}$ consists of a set of 2D image slices $\mathbf{X}_{i} = [\mathbf{x}_{i,1}, ..., \mathbf{x}_{i,D}]$ where $\mathbf{x}_{i,j}\in\mathbb{R}^{H \times W}$. 

Our proposed {\ours} framework consists of a mean-teacher network, $\mathcal{F}_t(\mathbf{X};\theta_t)$, and a student network, $\mathcal{F}_s(\mathbf{X};\theta_s)$. Inspired by recent work~\cite{laine2016temporal,tarvainen2017mean}, the optimization of these two networks can be achieved with an exponential moving average (EMA) which uses a weighted combination of the parameters of the student network and the parameters of the teacher network to update the latter. This strategy has been widely shown to improve training stability and the model's final performance. Motivated by this idea, our training strategy is divided into two steps. At each iteration, we first optimize the student network $\mathcal{F}_s$ by stochastic gradient descent. Then we update the teacher weights $\theta_{t}$ using an exponential moving average of the student weights $\theta_{s}$. 

The inputs to the two networks are perturbed versions of the same image. That is, given a volume input $\mathbf{X}_i$, we first add different perturbations (\ie, affine transformation and random crop) to generate two different images $\mathbf{X}_{i}^t$ and $\mathbf{X}_{i}^s$. We then feed $\mathcal{F}_t$ and $\mathcal{F}_s$ with these two corresponding augmented images to obtain two confidence score (probability) maps $\mathbf{Q}^t_i$ and $\mathbf{Q}^s_i$. 

Before we present our proposed {\ours} in detail, we first describe our base architecture below.

\subsection{Base Architecture}
\label{sec:model}

Our base architecture adopts V-Net~\cite{yu2019uncertainty} as the network backbone, which consists of an encoder network $e_t:\mathbb{R}^{H \times W \times D}\rightarrow \mathbb{R}^{H^{\prime} \times W^{\prime} \times D^{\prime} \times D_e}$ and a decoder network $d_t:\mathbb{R}^{H^{\prime} \times W^{\prime} \times D^{\prime} \times D_e}\rightarrow [0,1]^{H \times W \times D} \times [-1,1]^{H \times W \times D}$ for the teacher network, and similarly $e_s$, $d_s$ for the student network, \ie, $\mathcal{F}_t=d_t \circ e_t$ and $\mathcal{F}_s = d_s \circ e_s$. $H^{\prime}, W^{\prime}, D^{\prime}$ are the size of the hidden pattern and $D_e$ is the encoded feature dimension. Inspired by previous work on medical imaging segmentation \cite{xue2020shape,li2020shape}, we incorporate multi-task learning into $\mathcal{F}$ to jointly perform both classification and regression tasks.

Given input $\mathbf{X}_i$, the classification branch is designed to generate the probability map $\mathbf{Q}_i^s\in[0,1]^{H\times W\times D}$, and the regression branch is designed to predict the SDM $\mathbf{Q}_i^{s, \mathrm{sdm}}\in[-1,1]^{H\times W\times D}$. The design of the regression branch is simple yet effective, \ti{only} including the hyperbolic tangent function. This design brings two clear benefits: (1) we can eventually encode rich geometric structure information to improve segmentation accuracy, and (2) we can implicitly enforce continuity and smoothness terms for better segmentation maps. Similarly, we have outputs $\mathbf{Q}_i^{t}$ and $\mathbf{Q}_i^{t, \mathrm{sdm}}$ from the teacher network. 

\myparagraph{Supervised Loss $\mathcal{L}_{\mathrm{sup}}$}
For training on labeled data, we define the supervised loss as:
\begin{align}
    \mathcal{L}_{\mathrm{sup}} &= \frac{1}{N}\sum_{i=1}^{N}\mathcal{L}_{\mathrm{seg}}(\mathbf{Q}_i^{s},\mathbf{Y}_i) +  \frac{\alpha }{N}\sum_{i=1}^{N} \mathcal{L}_{\mathrm{mse}}(\mathbf{Q}^{s, \mathrm{sdm}}_{i}, \mathbf{Y}_i^{\mathrm{sdm}}) 
    \label{eq:supervised_loss}, 
\end{align}
where $\mathcal{L}_{\mathrm{seg}}$ denotes the segmentation loss (Dice and Cross-entropy)~\cite{yu2019uncertainty}, and $\mathcal{L}_{\mathrm{mse}}$ is the mean squared error loss. $\alpha$ is a hyperparameter. Note that the SDM loss~\cite{xue2020shape} is imposed as geometric constraints in training.

\begin{table*}[t]
\caption{Quantitative segmentation results on the LA dataset. The backbone network of all evaluated methods are V-Net.}
\label{tab:la}
\small
\centering
\resizebox{0.7\linewidth}{!}{
\begin{tabular}{c|c|c|c|c|c|c} 
\hline \hline
\multirow{2}{*}{Method} & \multicolumn{2}{c|}{\# \textbf{scans used}} & \multicolumn{4}{c}{\textbf{Metrics}}  \\ \cline{2-7}
    & Labeled
    & Unlabeled
    & Dice{[}\%{]}
    & Jaccard{[}\%{]} 
    & ASD{[}voxel{]} 
    & 95HD{[}voxel{]} \\ 
    \hline
    V-Net \cite{milletari2016v}       
    & 80             
    & 0                
    & 91.14        
    & 83.82           
    & 1.52           
    & 5.75            
    \\ \hline
    V-Net              
    & 16             
    & 0                
    & 86.03        
    & 76.06           
    & 3.51           
    & 14.26    
    \\ \hline  \hline
    DAN \cite{zhang2017deep}                     
    & 16             
    & 64               
    & 87.52        
    & 78.29           
    & 2.42           
    & 9.01            
    \\ \hline
    CPS \cite{chen2021semi} 
    & 16 
    & 64 
    & 87.87 
    & 78.61 
    & 2.16 
    & 12.87 
    \\ \hline
    MT \cite{tarvainen2017mean} 
    & 16 
    & 64 
    & 88.42 
    & 79.45 
    & 2.73 
    & 13.07 
    \\ \hline
    Entropy Mini \cite{vu2019advent} 
    & 16 
    & 64 
    & 88.45 
    & 79.51 
    & 3.72 
    & 14.14 
    \\ \hline
    UA-MT \cite{yu2019uncertainty}                   
    & 16             
    & 64               
    & 88.88        
    & 80.21           
    & 2.26           
    & 7.32            
    \\ \hline
    ICT \cite{verma2019interpolation} 
    & 16 
    & 64 
    & 89.02 
    & 80.34 
    & 1.97 
    & 10.38 
    \\ \hline
    SASSNet~\cite{li2020shape}          
    & 16             
    & 64               
    & 89.27        
    & 80.82           
    & 3.13           
    & 8.83            
    \\ \hline
    DTC~\cite{luo2020semi}          
    & 16             
    & 64               
    & 89.42        
    & 80.98           
    & 2.10           
    & 7.32            
    \\ \hline
    Chaitanya \etal \cite{chaitanya2020contrastive}          
    & 16             
    & 64               
    & {89.94}        
    & {81.82}           
    & {2.66}           
    & {7.23}            
    \\ \hline
    {\ours} (ours)    
    & 16             
    & 64               
    & \textbf{90.85}        
    & \textbf{83.80}           
    & \textbf{1.86}           
    & \textbf{6.03}            
    \\ \hline  \hline
    V-Net \cite{milletari2016v}                   
    & 8                                           
    & 0                                    
    & 79.99          
    & 68.12
    & 5.48           
    & 21.11           
    \\  \hline
    DAN \cite{zhang2017deep}                     
    & 8             
    & 72               
    & 75.11        
    & 63.47           
    & 3.57           
    & 19.04            
    \\ \hline
    CPS \cite{chen2021semi} 
    & 8 
    & 72 
    & 84.09 
    & 73.17 
    & 2.41 
    & 22.55
    \\ \hline
    MT \cite{tarvainen2017mean} 
    & 8 
    & 72
    & 84.24
    & 73.26 
    & 2.71
    & 19.41
    \\ \hline
    Entropy Mini \cite{vu2019advent} 
    & 8 
    & 72 
    & 85.90 
    & 75.60 
    & 2.74
    & 18.65 
    \\ \hline
    UA-MT \cite{yu2019uncertainty}                   
    & 8                                           
    & 72                                   
    & 84.25          
    & 73.48           
    & 3.36           
    & 13.84           
    \\ \hline
    ICT \cite{verma2019interpolation} 
    & 8 
    & 72 
    & 85.39 
    & 74.84 
    & 2.88 
    & 17.45 
    \\ \hline
    SASSNet~\cite{li2020shape}          
    & 8                                           
    & 72                                   
    & 86.81          
    & 76.92           
    & 3.94           
    & 12.54           
    \\ \hline
    DTC~\cite{luo2020semi}          
    & 8                                           
    & 72                                   
    & 87.49       
    & 78.03           
    & 2.37           
    & 9.06           
    \\ \hline
    Chaitanya \etal \cite{chaitanya2020contrastive}          
    & 8             
    & 72               
    & {84.95}        
    & {74.77}           
    & {3.70}           
    & {10.68}            
    \\ \hline
    {\ours} (ours)    
    & 8                                           
    & 72                                
    & \textbf{89.03} 
    & \textbf{80.34}  
    & \textbf{2.59}  
    & \textbf{8.34}   
    \\ \hline\hline
\end{tabular}}
\vspace{-10pt}
\end{table*}


\begin{figure*}[t]
	\centering
	\includegraphics[width=0.9\textwidth]{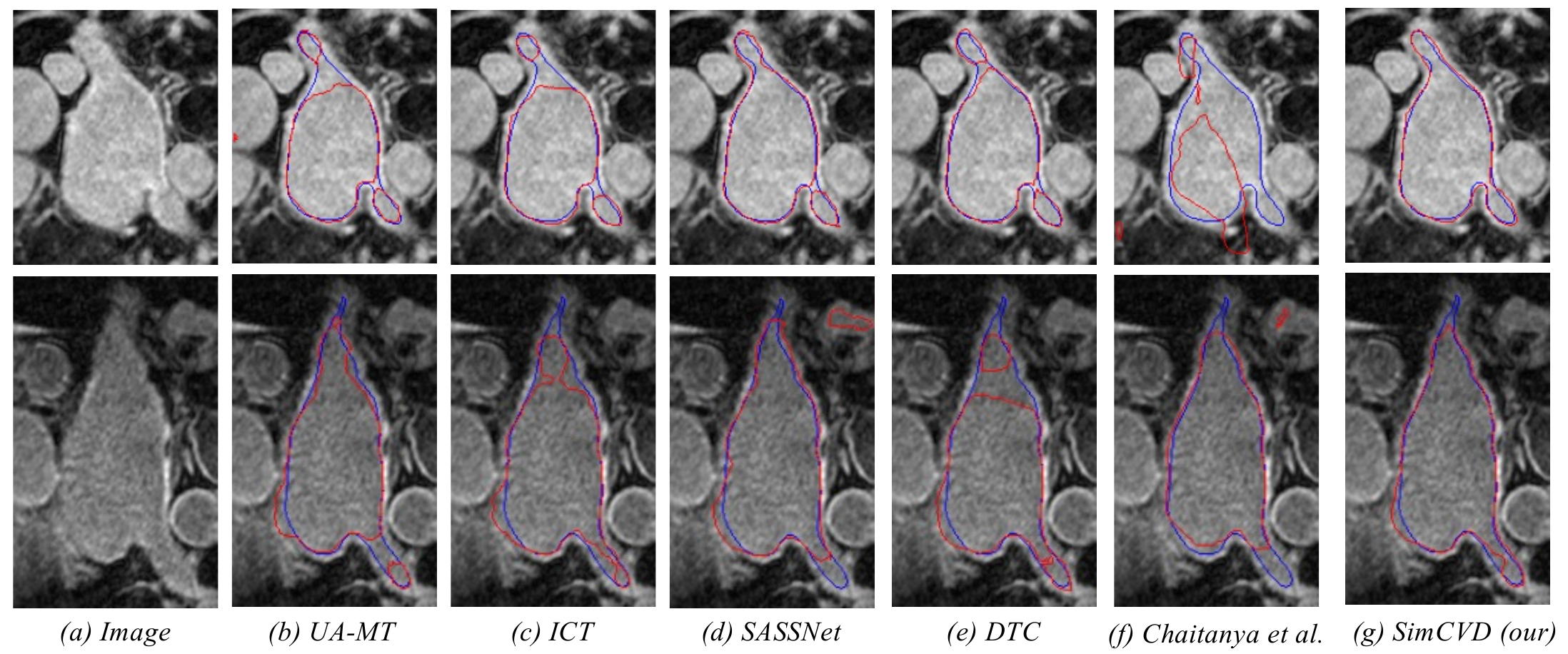}
    \vspace{-5pt}
	\caption{Visual comparisons with other methods on LA dataset. As observed, {\ours} achieves superior performance with more accurate borders and shapes. We train all the evaluated methods in the setting of 8 annotated images. \textcolor{red}{Red} and \textcolor{blue}{blue} denote the predictions and ground truths, respectively.}
    \vspace{-15pt}
	\label{fig:vis_la}
\end{figure*}

\subsection{Boundary-aware Contrastive Distillation}\label{sec:contrat}

Many prior methods distill knowledge merely in the shared prediction space by delivering the student network that matches the accuracy of the teacher network. However, this strategy is not robust for the following reasons: (1) the learned voxel-wise representations from the mean-teacher model are usually not robust due to the lack of geometric information; (2) the segmentation model still suffers from \textit{generalization} issues; and (3) the network performance needs to be further improved. Therefore, we propose to perform boundary-aware contrastive distillation to train our model for better segmentation accuracy. 

Our method differs from previous state-of-the-art methods in three aspects: (1) {\ours} imposes the global consistency in object boundary contours to capture more effective \ti{geometric information}; (2) previous methods follow the standard setting in considering the relations of local patches, while {\ours} aims to exploit correlations among all pairs of voxels to improve robustness; and (3) due to the computational cost, {\ours} does not use a large memory bank. {\ours} trains the contrastive objective as an auxiliary loss during the volume batch updates. To specify our voxel-wise contrastive distillation algorithm on unlabeled sets, we define two discrimination terms: boundary-aware contrastive loss and pair-wise distillation loss. 

\myparagraph{Boundary-aware Contrastive Loss $\mathcal{L}_{\mathrm{contrast}}$}
We describe our unsupervised boundary-aware contrastive objective as follows. Our key idea is to make use of ``boundary-aware'' knowledge by a contrastive learning objective that enforces the consistency of the predicted SDM outputs on the unlabeled set during training. The key ingredient to working with two views of input images is to apply \ti{dropout} as mask. Specifically, given the collection of an input volume $\mathbf{X}_i$, the student SDM $\mathbf{Q}_i^{s,\mathrm{sdm}}$, the teacher SDM $\mathbf{Q}_i^{t,\mathrm{sdm}}$, we first directly add them up to build two boundary-aware features: $\mathbf{Q}_i^{s, \mathrm{ba}} = \mathbf{X}_i + \mathbf{Q}_i^{s,\mathrm{sdm}}$ and $\mathbf{Q}_i^{t, \mathrm{ba}} = \mathbf{X}_i + \mathbf{Q}_i^{t,\mathrm{sdm}}$. Then, we feed them into the projection heads with two independent dropout masks $z_i^{s}, z_i^{t}$, and contrast~\textit{positives} and~\textit{negatives} by using the InfoNCE loss. We denote the same slice from the two boundary-aware features as \textit{positive}, and slices at different locations or from different inputs as \textit{negative}.

The boundary-aware features are created by adding the original 3D volume to the SDM because we want to fuse both the distance and the intensity information. Another way to achieve this is concatenation -- adding another dimension to the feature tensor -- requires a more complex projection head which is more prone to over-fitting. Thus, the projection head ${\mathcal{H}}\!:\!\mathbb{R}^{H \times W \times D} \!\!\rightarrow\! \mathbb{R}^{D_h \times D}$ encodes each 2D slice to a $D_h$-dimensional feature vector. The implementation is simple, which includes an alpha dropout \cite{klambauer2017self}, an adaptive average pooling, and a 3-layer multilayer perceptron (MLP). Here the MLP is designed to convert each 2D slice to a vector.

Denoting the output of the projection head as $\mf{H}_i^s =\mathcal{H}(\mathbf{Q}_i^{s, \mathrm{ba}}; z_i^s)$, $\mf{H}_i^t =\mathcal{H}(\mathbf{Q}_i^{t, \mathrm{ba}}; z_i^t)$, and the $jth$ row of $\mf{H}_i$ as $\mf{h}_{i,j}$, the InfoNCE loss~\cite{oord2018representation} is defined by:
\begin{equation}\label{eq:NCE}
\mathcal{L}(\mf{h}_{i,j}^{t}, \mf{h}_{i,j}^{s})=-\log\frac{\exp(\mf{h}_{i,j}^{t}\!\cdot\!\mf{h}_{i,j}^{s} / \tau)}{\sum_{k,l}\exp({\mf{h}_{i,j}^t\!\cdot\!\mf{h}_{k,l}^s /\tau})},\!\!
\vspace{-2pt}
\end{equation}
where $\tau$ is a temperature hyperparameter. The indices $k$ and $l$ in the denominator are randomly sampled from a mini-batch of images such that $B$ 2D slices are sampled in total. $i$ and $j$ denote the 3D image index and slice index, respectively. The $\mf{h}_{k,l}^s$'s in the denominator that are not $\mf{h}_{i,j}^s$ are called negative samples. Inspired by the recent success~\cite{chaitanya2020contrastive}, our boundary-aware contrastive loss is defined as:
\begin{equation}\label{eq:contrast_loss}
    \mathcal{L}_{\mathrm{contrast}}= \frac{1}{|\mathcal{N}^{+}|} \sum_{\forall(i,j) \in \mathcal{N}^{+}} [\mathcal{L}(\mf{h}_{i,j}^t,\mf{h}_{i,j}^s) + \mathcal{L}(\mf{h}_{i,j}^s,\mf{h}_{i,j}^t)],
    \vspace{-1pt}
\end{equation}
where $\mathcal{N}^{+}=\{ (i,j): i = N+1,\dots,N+M, j = 1,\dots, D \}$ denotes a collection of the positive 2D slice pairs. Note that the index $i$ in $\mathcal{N}^{+}$ is over all the unlabeled data, hence these data affect the training of $\mathcal{F}_s$ and $\mathcal{F}_t$.

\begin{table*}[t]
\caption{Quantitative segmentation results on the pancreas dataset. The backbone network of all evaluated methods are V-Net.}
\label{tab:pa}
\small
\centering
\resizebox{0.75\linewidth}{!}{
\begin{tabular}{c|c|c|c|c|c|c} 
\hline \hline
\multirow{2}{*}{Method} & \multicolumn{2}{c|}{\# \textbf{scans used}} & \multicolumn{4}{c}{\textbf{Metrics}}  \\ \cline{2-7}
    & Labeled
    & Unlabeled
    & Dice{[}\%{]}
    & Jaccard{[}\%{]} 
    & ASD{[}voxel{]} 
    & 95HD{[}voxel{]} \\ 
    \hline
    V-Net \cite{milletari2016v}       
    & 62             
    & 0                
    & 77.84        
    & 64.78           
    & 3.73           
    & 8.92            
    \\ \hline
    V-Net              
    & 12             
    & 0                
    & 62.42        
    & 48.06           
    & 4.77           
    & 22.34    
    \\ \hline  \hline
    MT \cite{tarvainen2017mean} 
    & 12             
    & 50               
    & 71.29 
    & 56.69
    & 2.82 
    & 16.31
    \\ \hline
    DAN \cite{zhang2017deep}                     
    & 12             
    & 50               
    & 68.67        
    & 53.97           
    & 3.07           
    & 15.78            
    \\ \hline
    CPS \cite{chen2021semi} 
    & 12             
    & 50               
    & 69.28 
    & 54.02 
    & 3.07 
    & 15.34 
    \\ \hline
    Entropy Mini \cite{vu2019advent} 
    & 12             
    & 50               
    & 69.33 
    & 54.32 
    & 2.92 
    & 15.29
    \\ \hline
    UA-MT \cite{yu2019uncertainty}
    & 12             
    & 50               
    & 72.43        
    & 57.91           
    & 4.25           
    & 11.01            
    \\ \hline
    ICT \cite{verma2019interpolation} 
    & 12             
    & 50               
    & 70.06      
    & 55.66           
    & 2.98
    & 13.05
    \\ \hline
    SASSNet~\cite{li2020shape}          
    & 12             
    & 50               
    & 70.47      
    & 55.74          
    & 4.26          
    & 10.95            
    \\ \hline
    DTC~\cite{luo2020semi}          
    & 12             
    & 50               
    & 74.07        
    & 60.17         
    & 2.61           
    & 10.35           
    \\ \hline
    Chaitanya \etal \cite{chaitanya2020contrastive}          
    & 12             
    & 50               
    & {70.79}        
    & {55.76}           
    & {6.08}           
    & {15.35}            
    \\ \hline
    {\ours} (ours)    
    & 12             
    & 50               
    & \textbf{75.39}        
    & \textbf{61.56}           
    & \textbf{2.33}           
    & \textbf{9.84}            
    \\ \hline \hline
\end{tabular}}
\vspace{-5pt}
\end{table*}


\begin{figure*}[t]
	\centering
	\includegraphics[width=0.9\textwidth]{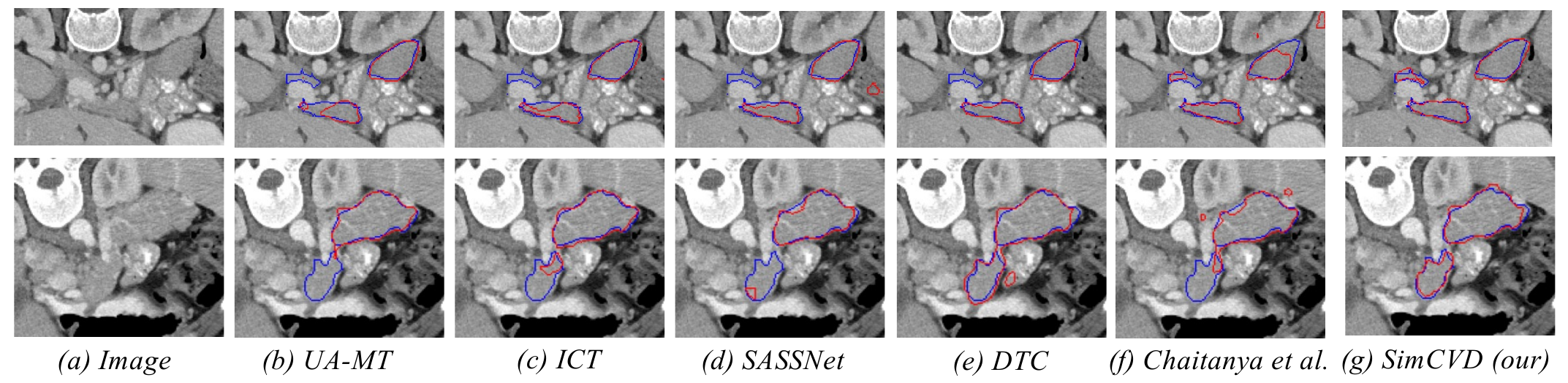}
    \vspace{-5pt}
	\caption{Visual comparisons with other methods on the pancreas dataset. We train all the evaluated methods in the setting of 12 annotated images. \textcolor{red}{Red} and \textcolor{blue}{blue} denotes the predictions and ground truths, respectively.}
    \vspace{-10pt}
	\label{fig:vis_pa}
\end{figure*}

\myparagraph{Pair-wise Distillation Loss $\mathcal{L}_{\mathrm{pd}}$}
On one hand, boundary-aware contrastive objectives uncover distinctive global boundary-aware representations that benefit the training of downstream tasks, \eg~object classification, when limited labeled data is available. On the other hand, dense predictive tasks,~\eg~semantic segmentation, may require more discriminative spatial representations. As complementary to boundary-aware contrastive objectives, a promising local pair-wise strategy is vital for the medical image segmentation tasks. With this insight, we propose to perform voxel-to-voxel pair-wise distillation to explicitly explore structural relationships between voxel samples to improve spatial labeling consistency.

In our implementation, we enforce such a constraint on the hidden patterns from the encoders $e_t$ and $e_s$. Specifically, let $\mf{V}_i^t\in \mathbb{R}^{H^{\prime} W^{\prime} D^{\prime} \times D_e}$ and $\mf{V}_i^s\in \mathbb{R}^{H^{\prime} W^{\prime} D^{\prime} \times D_e}$ be the first-3-dimension-flattened hidden patterns of $e_t(\mathbf{X}_i)$ and $e_s(\mathbf{X}_i)$ respectively, and $\mf{v}_{i,j}$ be the $jth$ row of $\mf{V}_{i}$. The pair-wise distillation loss is defined as:
\begin{equation}\label{eq:distillation_loss}
    \mathcal{L}_{\mathrm{pd}} = -\frac{1}{M}\sum_{i=N+1}^{N+M} \sum_{j=1}^{H^{\prime} W^{\prime} D^{\prime}} 
    \log \frac{\exp (s(\mathbf{v}_{i,j}^s, \mathbf{v}_{i,j}^t))}{\sum_k \exp (s(\mathbf{v}_{i,j}^s, \mathbf{v}_{i,k}^t))},
    \vspace{-3pt}
\end{equation}
where $s(\mathbf{v}_1, \mathbf{v}_2) = \frac{\mathbf{v}_1 \cdot \mathbf{v}_2}{\| \mathbf{v}_1 \| \| \mathbf{v}_2 \|}$ measures the cosine of the angle between two $\mathbf{v}$'s as their similarity. Note again that this loss also involves all the unlabeled data.

\myparagraph{Consistency Loss $\mathcal{L}_{\mathrm{con}}$}
Inspired by recent work \cite{laine2016temporal,tarvainen2017mean}, consistency is designed to further encourage training stability and performance improvements on the unlabeled set. In our implementations, we first perform different perturbation operations on the unlabeled input volume $\mf{X}_{i}$, \ie, adding noise~$\eta_i$, and then define the consistency loss as:
\begin{equation}\label{eq:mean-tea}
    \mathcal{L}_{\mathrm{con}}=\frac{1}{M} \sum_{i=N+1}^{N+M}\mathcal{L}_{\mathrm{mse}}\left(\mathcal{F}_s\left(\mf{X}_{i}^s + \eta_i^s\right), \mathcal{F}_t\left(\mf{X}_{i}^{t} + \eta_i^t\right)\right).
    \vspace{-3pt}
\end{equation}

\myparagraph{Overall Training Objective} 
{\ours} is a general semi-supervised framework for combining contrastive distillation with geometric constraints. In our experiments, we train {\ours} with two objective functions --- a supervised objective and an unsupervised objective. For the labeled data, we define the supervised loss in Section~\ref{sec:model}. For the unlabeled data, the unsupervised training objective consists of the boundary-aware contrastive loss, pair-wise distillation loss, and consistency loss in Section \ref{sec:contrat}. The overall loss function is:
\begin{equation}\label{eq:overall_obj}
    \mathcal{L} = \mathcal{L}_{\mathrm{sup}} + \lambda \mathcal{L}_{\mathrm{contrast}} + \beta \mathcal{L}_{\mathrm{pd}} + \gamma \mathcal{L}_{\mathrm{con}},
    \vspace{-3pt}
\end{equation}
where $\lambda,\beta,\gamma$ are hyperparameters that balance each term.

\section{Experimental Setup}
\label{sec:experiments}

\subsection{Dataset and Pre-processing}
We evaluated our approach on two popular benchmark datasets: the Left Atrium (LA) MR dataset from the Atrial Segmentation Challenge\footnote{http://atriaseg2018.cardiacatlas.org/}, and the NIH pancreas CT dataset \cite{roth2016data}. For the Left Atrium dataset, it comprises 100 3D gadolinium-enhanced MR imaging scans (GE-MRIs) with expert annotations, with an isotropic resolution of $0.625\times 0.625 \times 0.625 \text{mm}^3$. Following the experimental setting in \cite{yu2019uncertainty}, we use 80 scans for training, and 20 scans for evaluation. We employ the same pre-processing methods by cropping all the scans at the heart region and normalizing the intensities to zero mean and unit variance. All the training sub-volumes are augmented by random cropping to~$112\times 112\times 80 \text{mm}^3$. For the pancreas dataset, it contains 82 contrast-enhanced abdominal CT scans. Following the experimental settings in \cite{luo2020semi}, we randomly select 62 scans for training, and 20 scans for evaluation. In the pre-processing, we first truncate the intensities of the CT images into the window [$-125$, $275$] HU~\cite{zhou2019prior}, and then resample all the data into a fixed isotropic resolution of $1.0\times1.0\times1.0 \text{mm}^3$. Finally, we crop all the scans centering at the pancreas region, and normalize the intensities to zero mean and unit variance. All the training sub-volumes are augmented by random cropping to~$96\times 96\times 96 \text{mm}^3$.  In this study, we compare all the
methods on LA and the pancreas dataset with respect to 20\% labeled ratio. To emphasize the effectiveness of {\ours}, we further validate all the methods with respect to 10\% labeled ratio on LA dataset.

\subsection{Implementation Details}
In this study, all evaluated methods are implemented in PyTorch, and trained for $6000$ iterations on an NVIDIA 1080Ti GPU with a batch size of $4$. For data augmentation, we use standard data augmentation techniques (\ie, random rotation, flipping, and cropping). We set the hyper-parameters $\alpha$,\,$\lambda$,\,$\beta$,\,$\gamma$,\,$\tau$ as $0.1$,\,$0.5$,\,$0.1$,\,$0.1$,\,$0.5$, respectively. For the projection head, we set $p = 0.1$ in the \ti{AlphaDropout} layer, and output size $128\times128$ for \ti{AdaptiveAvgPool2d}. We use SGD optimizer with a momentum of \!$0.9$ and a weight decay of \!$0.0005$ to optimize network parameters. The initial learning rate is set as $0.01$ and divided by $10$ every $3000$ iterations. For EMA updates, we follow the experimental setting in \cite{yu2019uncertainty}, where the EMA decay rate $\alpha$ is set to $0.999$. We use the time-dependent Gaussian warming-up function \,$\Psi_{\text{con}}(t)\!=\!\exp{\left(-5\left(1-t / t_{\max}\right)^{2}\right)}$\, to ramp up parameters, where $t$ and $t_{\max}$ denote the current and the maximum training step, respectively. For fairness, we do not adopt any post-processing step.

In the testing stage, we adopt four metrics to evaluate the segmentation performance: Dice coefficient (Dice), Jaccard Index (Jaccard), 95\% Hausdorff Distance (95HD), and Average Symmetric Surface Distance (ASD). Following~\cite{yu2019uncertainty,luo2020semi,ssl4mis2020}, we adopt a sliding window strategy, which uses a stride with $18\times18\times4$  for the LA and $16\times16\times16$ for the pancreas.
\section{Results}

\subsection{Experiments: Left Atrium}

We compare {\ours} with published results from previous state-of-the-art semi-supervised segmentation methods, including V-Net \cite{milletari2016v}, MT \cite{tarvainen2017mean}, DAN \cite{zhang2017deep}, CPS \cite{chen2021semi}, Entropy Mini \cite{vu2019advent}, UA-MT \cite{yu2019uncertainty}, ICT \cite{verma2019interpolation}, SASSNet \cite{li2020shape}, DCT \cite{luo2020semi}, and Chaitanya \etal \cite{chaitanya2020contrastive} on the LA dataset in two labeled ratio settings (\ie, 10\% and 20\%). 

The quantitative results on the LA dataset are shown in Table \ref{tab:la}. {\ours} substantially improved the segmentation accuracy in both 10\% and 20\% labeled cases. The results are visualized in Fig \ref{fig:vis_la}. Specifically, in the setting of 20\% labeled ratio, our proposed {\ours} raises the previous best average results from 89.94\% to 90.85\% and from 81.82\% to 83.80\% in terms of Dice and Jaccard, even achieving comparable performance to the fully supervised baseline. Using the 10\% labeled ratio, {\ours} further advances the state-of-the-art results from 87.49\% to 89.03\% in Dice. The gains in Jaccard, ASD, and 95HD are also substantial, achieving 80.34\%, 2.59, and 8.34, respectively. This suggests that: (1) taking voxel samples with a contrastive objective yields better voxel embeddings; (2) incorporating pair-wise spatial labeling consistency can boost the performance by accessing more structural knowledge; and (3) utilizing a geometric constraint (\ie, SDM) is capable of helping identify more accurate boundaries. Leveraging all these aspects, we can observe consistent performance gains.

\subsection{Experiments: Pancreas}

To further evaluate the effectiveness of {\ours}, we compare our model on the pancreas CT dataset. Experimental results on the pancreas CT dataset are summarized in Table \ref{tab:pa}. We observe that our model consistently outperforms all previous methods, achieving up to 6.72\% absolute improvements in Dice. As shown in Figs. \ref{fig:vis_la} and \ref{fig:vis_pa}, our method is capable of predicting high-quality object segmentation, considering the fact that the improvement in such a setting is difficult. This demonstrates: (1) the necessity of comprehensively considering both boundary-aware contrast and pair-wise distillation; and (2) the efficacy of global shape information. Compared to the previous strong models, our approach achieves large improvements on all the datasets, demonstrating its effectiveness.

\begin{table}[t]
\caption{Ablation on {\bf (a)} model component: {\ours}  w/o SDM; {\ours}  w/o adaptive max pooling; {\bf (b)} loss formulation: {\ours} w/o $\mathcal{L}_\mathrm{contrast}$; {\ours} w/o $\mathcal{L}_\mathrm{pd}$; {\ours}  w/o $\mathcal{L}_\mathrm{sdm}$, compared to the baseline and our proposed {\ours}.}
\vspace{-5pt}
\label{tab:component_ablation}
\resizebox{\linewidth}{!}{
\begin{tabular}{c|l|c c c c c} 
\hline \hline 
    & \multirow{2}{*}{Method} & \multicolumn{4}{c}{\textbf{Metrics}} & \multirow{2}{*}{p-value (vs. SimCVD, {[}\%{]})}
    \\ 
    \cline{3-6}
    & & Dice{[}\%{]}
    & Jaccard{[}\%{]} 
    & ASD{[}voxel{]} 
    & 95HD{[}voxel{]}
    \\ \hline
    & Baseline (UA-MT)
    & 84.24
    & 73.26 
    & 2.71
    & 19.41
    & 0.019
    \\ \hline
    \multirow{2}{*}{\bf (a)} 
    & {\ours}  w/o SDM
    & 88.24 
    & 79.07 
    & 4.19 
    & 11.43
    & 2.47
    \\
    & {\ours} 
    w/ Adaptive Max Pooling
    & 88.32 
    & 79.26 
    & 3.79
    & 14.69
    & 0.33
    \\ \hline
    \multirow{3}{*}{\bf (b)}
    & {\ours} 
    w/o $\mathcal{L}_\mathrm{contrast} + \mathcal{L}_\mathrm{pd}$
    & 84.97
    & 74.49
    & 6.13
    & 19.98
    & 0.018
    \\
    & {\ours}  w/o $\mathcal{L}_\mathrm{contrast}$
    & 85.13 
    & 74.57
    & 5.97 
    & 16.61
    & 2.9e-6
    \\
    & {\ours}  
    w/o $\mathcal{L}_\mathrm{pd}$ 
    & 88.11
    & 78.89
    & 2.89
    & 12.58
    & 2.02
    \\
    & {\ours} 
    w/o $\mathcal{L}_\mathrm{sdm}$
    & 88.85
    & 80.03 
    & 2.71
    & 9.02
    & 5.14
    \\ \hline
	& {\ours}
    & \textbf{89.03} 
    & \textbf{80.34}  
    & \textbf{2.59}  
    & \textbf{8.34}
    & -
    \\ \hline\hline
\end{tabular}
}
\vspace{-5pt}
\end{table}


\section{Ablation Study}
\label{sec:ablation}

In this section, we conduct extensive studies to better understand {\ours}. We justify the inner working of {\ours} from two perspectives: (1) boundary-aware contrastive distillation (Section \ref{subsec:component}), and (2) the projection head (Section \ref{subsec:projection}). In these studies, we evaluate our proposed method on the LA dataset with 10\% labeled ratio (8 labeled and 72 unlabeled).

\subsection{Analysis on Boundary-aware Contrastive Distillation}
\label{subsec:component}

\myparagraph{Ablation on Model Component}
In the model formulation, our motivation is to advance state-of-the-art voxel-wise representations by capturing the geometric and semantic information in 3D space. Rather than transferring knowledge across confidence score maps directly, our {\ours} distills ``boundary-aware'' knowledge from the teacher network. To validate the idea of boundary-aware contrastive distillation, we compare {\ours} to an ablative baseline (\ie, {\ours}  w/o SDM). Table \ref{tab:component_ablation} {\bf(a)} compares each component of {\ours} in the 10\% labeled setting. First, we observe that removing the SDMs in training hurts the segmentation performance by $\!-0.79\%$, $\!-1.27\%$, $\!-1.6$, and $\!-3.09$ absolute differences in terms of Dice, Jaccard, ASD, and 95HD. This confirms our intuition that the learned boundary-aware representations provide a good prior for improving segmentation accuracy. We also find that using adaptive max pooling strategy (\ie, {\ours}  w/ adaptive max pooling) largely degrades the segmentation performance. Our segmentation results demonstrate that {{\ours}} is an effective approach, outperforming the best previous method with $\!+0.71\%$, $\!+1.08\%$, $\!+1.20$, and $\!+6.35$ absolute differences in terms of Dice, Jaccard, ASD, and 95HD. We hypothesize that it is because ``w/ adaptive max pooling'' leads to information loss during training.


\begin{table}[t]
\caption{Ablation on dropout rates $p$ and pooling size.}
\vspace{-5pt}
\label{tab:projection_ablation}
\resizebox{\linewidth}{!}{
\begin{tabular}{c|l|c c c c} 
\hline \hline 
    & \multirow{2}{*}{Method} & \multicolumn{4}{c}{\textbf{Metrics}} \\ \cline{3-6}
    & & Dice{[}\%{]}
    & Jaccard{[}\%{]} 
    & ASD{[}voxel{]} 
    & 95HD{[}voxel{]} 
    \\ \hline
    \multirow{5}{*}{Dropout} 
    & $p=0.0$
    & 87.69
    & 78.23
    & 2.49
    & 11.03 
    \\
    & $p=0.01$
    & 87.98
    & 78.69 
    & 3.08
    & 11.17
    \\
    & $p=0.02$
    & 87.99
    & 78.70
    & 2.60
    & 9.03
    \\
    & $p=0.05$
    & 88.01
    & 78.71
    & 2.78
    & 10.60
    \\
    & $p=0.1$
    & {89.03} 
    & {80.34}  
    & {2.59}  
    & {8.34}
    \\
    & $p=0.2$
    & {88.10} 
    & {78.86}  
    & {3.28}  
    & {12.70}
    \\
    & $p=0.5$
    & 86.67
    & 76.59
    & 4.20
    & 14.89 
    \\ \hline
    \multirow{5}{*}{Pooling Size}
    & $16\times 16$
    & 87.04
    & 77.22
    & 3.69
    & 14.28 
    \\
    & $32\times 32$
    & 87.53
    & 77.98
    & 3.13
    & 11.56
    \\
    & $64\times 64$
    & 88.37
    & 79.27
    & 2.75
    & 8.84 
    \\
    & $128\times 128$
    & {89.03} 
    & {80.34}  
    & {2.59}  
    & {8.34}  
    \\
    & $256\times 256$
    & 87.99 
    & 78.71
    & 2.82
    & 9.97
    \\ \hline\hline
\end{tabular}
}
\vspace{-10pt}
\end{table}


\myparagraph{Ablation on Loss Formulation}
In the loss formulation, our main idea is to pull closer similar (\ti{positive}) pairs upon the same threshold, while pushing apart dissimilar (\ti{negative}) pairs. Our learning objective is designed to jointly exploit effective correlations in the \ti{prediction} and \ti{feature} space in an informative way. To evaluate the effectiveness of each objective term, we conduct ablation studies by removing each term separately. As shown in Table \ref{tab:component_ablation} {\bf(b)}, we observe that {\ours} outperforms the ablative baseline ``{\ours} w/o $\mathcal{L}_\mathrm{contrast}$'' by a large margin and achieves a $\!+3.90\%\!$ increase in Dice. It clearly demonstrates it can effectively capture global context structure and local cues of 3D shapes. Next, we study whether enforcing similarity constraints in the \ti{feature} space is effective enough to exploit structured knowledge in practice. We find that, under the 10\% labeled setting, removing $\mathcal{L}_\mathrm{pd}$ leads to a performance drop, and the accuracy measures decreases by $\!-0.92\%\!$ and $\!-1.45\%\!$ in Dice and Jaccard, respectively. Our qualitative results confirm that \textit{pair-wise distillation} is effective for {\ours} in improving the network performance. When we remove $\mathcal{L}_\mathrm{sdm}$, the network performance does not drop significantly. We speculate that our boundary-aware contrastive distillation framework is capable of eliciting ``boundary-aware'' knowledge from the teacher model with high accuracy. This further highlights the effectiveness of our proposed {\ours}. To further verify the robustness of {\ours}, we perform pair-wise t-test between {\ours} and the other methods, using the per-case test Dice score as samples. The null hypothesis states that the test scores come from the same distribution, and thus the methods do not differ; whereas the alternative hypothesis is that {\ours} yields higher Dice score. For all the alternative methods in the Table below, the p-values are close to or less than $0.05$. Small p-values indicates that we are confident in rejecting the null hypothesis and conclude that {\ours} indeed outperforms the baseline models.

\subsection{Analysis on Projection Head}
\label{subsec:projection}
To further understand how different aspects of our projection head contribute to the superior model performance, we conduct extensive experiments and discuss our findings below.

\myparagraph{How to Interpret Dropout?}
Our experimental results have shown that {{\ours}} is an effective approach. In the following, we aim to answer two questions. First, how can we interpret {\ours}'s \ti{dropout} training strategy? Can we view \ti{dropout} as a form of data augmentation? Second, is it capable of exploiting additional informative cues in practice?

First, we examine whether removing \ti{dropout} during training can achieve comparable performance. Table \ref{tab:projection_ablation} shows the ablation result of our \ti{dropout} on LA. As shown in Table \ref{tab:projection_ablation}, we observe that using \ti{dropout} achieves a much better result on LA dataset. Compared to the setting $p\!=\!0.1$, we find that ``no dropout'' ($p\!=\!0$) leads to a dramatic performance degradation by $\!-1.34\%$, $\!-2.11\%$, $\!-1.71$, $\!-2.69$ absolute differences in terms of Dice, Jaccard, ASD, and 95HD, respectively. While in the case of $p\!=\!0.5$, it also significantly hurts the network performance. On the other hand, we observe slight improvements on the other $p$ settings, compared to ``no dropout'', but eventually underperform {\ours}. This clearly demonstrates the superiority of our \ti{dropout} strategy to learn better representations with respect to different pairs of augmented images. We speculate that adding \ti{dropout} can be interpreted as a minimal form of data augmentation, in which the positive pair takes two views of the same images, and their representations make a clear difference in dropout masks.

\myparagraph{Effect of Augmentation Techniques}
To further examine our hypothesis, we compare common data augmentation techniques (\ie, local shuffle pixel, non-linear transformation, in-painting, out-painting) in Table \ref{tab:data_aug}. As is shown, the quantitative results reveal interesting behavior of different data augmentation: adding more data augmentation does not further contribute to the good model performance. We note that, somewhat surprisingly, it hurts the final prediction performance, and none of them outperforms the basic \ti{dropout} mask. This suggests that by including these data augmentation techniques, it is possible to introduce additional noise during training, which leads to the representation collapse.


\begin{table}[t]
\caption{
    Ablation on different data augmentations on the LA dataset. All of them include the dropout masks ($p=0.1$).}
\label{tab:data_aug}
\vspace{-5pt}
\resizebox{\linewidth}{!}{
\begin{tabular}{l|c c c c} 
\hline \hline 
    \multirow{2}{*}{Method} & \multicolumn{4}{c}{\textbf{Metrics}} \\ \cline{2-5}
    & Dice{[}\%{]}
    & Jaccard{[}\%{]} 
    & ASD{[}voxel{]} 
    & 95HD{[}voxel{]} 
    \\ \hline
    {\ours}  
    & {89.03} 
    & {80.34}  
    & {2.59}  
    & {8.34}   
    \\ \hline
    \quad + Local Shuffle Pixel
    & 88.15 
    & 78.97
    & 1.96
    & 8.66
    \\
    \quad + Non-linear Intensity Transformation
    & 88.02
    & 78.80
    & 2.68 
    & 10.29
    \\
    \quad + In-painting
    & 88.37
    & 79.26
    & 2.84
    & 10.97
    \\
    \quad + Out-painting
    & 88.24
    & 79.07
    & 2.58
    & 10.62
    \\ \hline\hline
\end{tabular}
}
\vspace{-10pt}
\end{table}


\myparagraph{Effect of Pooling Size}
In Table \ref{tab:component_ablation}, we demonstrate the network improvements from using adaptive mean pooling instead of adaptive max pooling. We investigate the effects of different pooling sizes in Table \ref{tab:projection_ablation}. Empirically, we observe that using a larger pooling size clearly improves performance consistently. However, we find that the results can not be improved further by increasing the pooling size to $256$. In our implementation, we set the pooling size as $128$.
\section{Conclusion}
\label{section:conclusion}

In this work, we propose {\ours}, a simple contrastive distillation learning framework, which largely advances state-of-the-art voxel-wise representation learning on medical segmentation tasks. Specifically, we present an unsupervised training strategy, which takes two views of an input volume and predicts their signed distance maps of their object boundaries in a contrastive objective, with only two different dropout masks. We further conduct extensive analyses to understand the state-of-the-art performance of our approach, and demonstrate the importance of learning distinct boundary-aware representations and using dropout as the minimal data augmentation technique. We also propose to perform structural distillation by distilling pair-wise similarities, which achieves good performance improvements. Our experimental results show that {\ours} obtained new state-of-the-art results on two benchmarks in an extreme few-annotation setting. 

We believe that our unsupervised training framework provides a new perspective on data augmentation along with unlabeled 3D medical data. We also plan to extend our method to solve multi-class medical image segmentation tasks.

\bibliographystyle{ieeetr}
\bibliography{tmi}

\begin{thebibliography}{10}

\bibitem{staib1996model}
L.~H. Staib and J.~S. Duncan, ``Model-based deformable surface finding for
  medical images,'' {\em IEEE Transactions on Medical Imaging}, vol.~15, no.~5,
  pp.~720--731, 1996.

\bibitem{yang2004neighbor}
J.~Yang, L.~H. Staib, and J.~S. Duncan, ``Neighbor-constrained segmentation
  with level set based 3-d deformable models,'' {\em IEEE Transactions on
  Medical Imaging}, vol.~23, no.~8, pp.~940--948, 2004.

\bibitem{yang20043d}
J.~Yang and J.~S. Duncan, ``{3D} image segmentation of deformable objects with
  joint shape-intensity prior models using level sets,'' {\em Medical Image
  Analysis}, vol.~8, no.~3, pp.~285--294, 2004.

\bibitem{chakraborty1996deformable}
A.~Chakraborty, L.~H. Staib, and J.~S. Duncan, ``Deformable boundary finding in
  medical images by integrating gradient and region information,'' {\em IEEE
  Transactions on Medical Imaging}, vol.~15, no.~6, pp.~859--870, 1996.

\bibitem{staib1992boundary}
L.~H. Staib and J.~S. Duncan, ``Boundary finding with parametrically deformable
  models,'' {\em IEEE Transactions on Pattern Analysis and Machine
  Intelligence}, vol.~14, no.~11, pp.~1061--1075, 1992.

\bibitem{ronneberger2015u}
O.~Ronneberger, P.~Fischer, and T.~Brox, ``U-net: Convolutional networks for
  biomedical image segmentation,'' in {\em MICCAI}, pp.~234--241, Springer,
  2015.

\bibitem{milletari2016v}
F.~Milletari, N.~Navab, and S.-A. Ahmadi, ``V-net: Fully convolutional neural
  networks for volumetric medical image segmentation,'' in {\em 3DV},
  pp.~565--571, IEEE, 2016.

\bibitem{bai2017semi}
W.~Bai, O.~Oktay, M.~Sinclair, H.~Suzuki, M.~Rajchl, G.~Tarroni, B.~Glocker,
  A.~King, P.~M. Matthews, and D.~Rueckert, ``Semi-supervised learning for
  network-based cardiac {MR} image segmentation,'' in {\em MICCAI},
  pp.~253--260, Springer, 2017.

\bibitem{ganaye2018semi}
P.-A. Ganaye, M.~Sdika, and H.~Benoit-Cattin, ``Semi-supervised learning for
  segmentation under semantic constraint,'' in {\em MICCAI}, pp.~595--602,
  Springer, 2018.

\bibitem{you2020unsupervised}
C.~You, J.~Yang, J.~Chapiro, and J.~S. Duncan, ``Unsupervised wasserstein
  distance guided domain adaptation for {3D} multi-domain liver segmentation,''
  in {\em Interpretable and Annotation-Efficient Learning for Medical Image
  Computing}, pp.~155--163, Springer, 2020.

\bibitem{wang2020deep}
Y.~Wang, X.~Wei, F.~Liu, J.~Chen, Y.~Zhou, W.~Shen, E.~K. Fishman, and A.~L.
  Yuille, ``Deep distance transform for tubular structure segmentation in ct
  scans,'' in {\em CVPR}, pp.~3833--3842, 2020.

\bibitem{xue2020shape}
Y.~Xue, H.~Tang, Z.~Qiao, G.~Gong, Y.~Yin, Z.~Qian, C.~Huang, W.~Fan, and
  X.~Huang, ``Shape-aware organ segmentation by predicting signed distance
  maps,'' in {\em AAAI}, 2020.

\bibitem{li2020transformation}
X.~Li, L.~Yu, H.~Chen, C.-W. Fu, L.~Xing, and P.-A. Heng,
  ``Transformation-consistent self-ensembling model for semisupervised medical
  image segmentation,'' {\em IEEE Transactions on Neural Networks and Learning
  Systems}, 2020.

\bibitem{laine2016temporal}
S.~Laine and T.~Aila, ``Temporal ensembling for semi-supervised learning,''
  {\em arXiv preprint arXiv:1610.02242}, 2016.

\bibitem{zhang2017deep}
Y.~Zhang, L.~Yang, J.~Chen, M.~Fredericksen, D.~P. Hughes, and D.~Z. Chen,
  ``Deep adversarial networks for biomedical image segmentation utilizing
  unannotated images,'' in {\em MICCAI}, pp.~408--416, Springer, 2017.

\bibitem{li2018semi}
X.~Li, L.~Yu, H.~Chen, C.-W. Fu, and P.-A. Heng, ``Semi-supervised skin lesion
  segmentation via transformation consistent self-ensembling model,'' {\em
  arXiv preprint arXiv:1808.03887}, 2018.

\bibitem{nie2018asdnet}
D.~Nie, Y.~Gao, L.~Wang, and D.~Shen, ``Asdnet: Attention based semi-supervised
  deep networks for medical image segmentation,'' in {\em MICCAI},
  pp.~370--378, Springer, 2018.

\bibitem{qiao2018deep}
S.~Qiao, W.~Shen, Z.~Zhang, B.~Wang, and A.~Yuille, ``Deep co-training for
  semi-supervised image recognition,'' in {\em ECCV}, pp.~135--152, 2018.

\bibitem{bortsova2019semi}
G.~Bortsova, F.~Dubost, L.~Hogeweg, I.~Katramados, and M.~de~Bruijne,
  ``Semi-supervised medical image segmentation via learning consistency under
  transformations,'' in {\em MICCAI}, pp.~810--818, Springer, 2019.

\bibitem{li2021assessing}
K.~Li, W.~Zhou, H.~Li, and M.~A. Anastasio, ``Assessing the impact of deep
  neural network-based image denoising on binary signal detection tasks,'' {\em
  IEEE transactions on medical imaging}, vol.~40, no.~9, pp.~2295--2305, 2021.

\bibitem{yu2019uncertainty}
L.~Yu, S.~Wang, X.~Li, C.-W. Fu, and P.-A. Heng, ``Uncertainty-aware
  self-ensembling model for semi-supervised {3D} left atrium segmentation,'' in
  {\em MICCAI}, pp.~605--613, Springer, 2019.

\bibitem{li2020shape}
S.~Li, C.~Zhang, and X.~He, ``Shape-aware semi-supervised {3D} semantic
  segmentation for medical images,'' in {\em MICCAI}, pp.~552--561, Springer,
  2020.

\bibitem{zhu2020rubik}
J.~Zhu, Y.~Li, Y.~Hu, K.~Ma, S.~K. Zhou, and Y.~Zheng, ``Rubik’s cube+: A
  self-supervised feature learning framework for {3D} medical image analysis,''
  {\em Medical Image Analysis}, p.~101746, 2020.

\bibitem{luo2020semi}
X.~Luo, J.~Chen, T.~Song, and G.~Wang, ``Semi-supervised medical image
  segmentation through dual-task consistency,'' in {\em AAAI}, 2020.

\bibitem{chaitanya2020contrastive}
K.~Chaitanya, E.~Erdil, N.~Karani, and E.~Konukoglu, ``Contrastive learning of
  global and local features for medical image segmentation with limited
  annotations,'' in {\em NeurIPS}, 2020.

\bibitem{you2021momentum}
C.~You, R.~Zhao, L.~Staib, and J.~S. Duncan, ``Momentum contrastive voxel-wise
  representation learning for semi-supervised volumetric medical image
  segmentation,'' {\em arXiv preprint arXiv:2105.07059}, 2021.

\bibitem{chen2020simple}
T.~Chen, S.~Kornblith, M.~Norouzi, and G.~Hinton, ``A simple framework for
  contrastive learning of visual representations,'' in {\em ICML}, 2020.

\bibitem{hjelm2018learning}
R.~D. Hjelm, A.~Fedorov, S.~Lavoie-Marchildon, K.~Grewal, P.~Bachman,
  A.~Trischler, and Y.~Bengio, ``Learning deep representations by mutual
  information estimation and maximization,'' in {\em ICLR}, 2019.

\bibitem{bai2019self}
W.~Bai, C.~Chen, G.~Tarroni, J.~Duan, F.~Guitton, S.~E. Petersen, Y.~Guo, P.~M.
  Matthews, and D.~Rueckert, ``Self-supervised learning for cardiac {MR} image
  segmentation by anatomical position prediction,'' in {\em MICCAI},
  pp.~541--549, Springer, 2019.

\bibitem{peng2021self}
J.~Peng, P.~Wang, C.~Desrosiers, and M.~Pedersoli, ``Self-paced contrastive
  learning for semi-supervised medical image segmentation with meta-labels,''
  in {\em NeurIPS}, 2021.

\bibitem{yang2019snapshot}
C.~Yang, L.~Xie, C.~Su, and A.~L. Yuille, ``Snapshot distillation:
  Teacher-student optimization in one generation,'' in {\em CVPR},
  pp.~2859--2868, 2019.

\bibitem{zhuang2020deep}
J.~Zhuang, J.~Cai, R.~Wang, J.~Zhang, and W.-S. Zheng, ``Deep knn for medical
  image classification,'' in {\em MICCAI}, pp.~127--136, Springer, 2020.

\bibitem{tarvainen2017mean}
A.~Tarvainen and H.~Valpola, ``Mean teachers are better role models:
  Weight-averaged consistency targets improve semi-supervised deep learning
  results,'' in {\em NeurIPS}, pp.~1195--1204, 2017.

\bibitem{srivastava2014dropout}
N.~Srivastava, G.~Hinton, A.~Krizhevsky, I.~Sutskever, and R.~Salakhutdinov,
  ``Dropout: a simple way to prevent neural networks from overfitting,'' {\em
  The Journal of Machine Learning Research}, vol.~15, no.~1, pp.~1929--1958,
  2014.

\bibitem{perera2015motion}
S.~Perera, N.~Barnes, X.~He, S.~Izadi, P.~Kohli, and B.~Glocker, ``Motion
  segmentation of truncated signed distance function based volumetric
  surfaces,'' in {\em WACV}, pp.~1046--1053, IEEE, 2015.

\bibitem{dangi2019distance}
S.~Dangi, C.~A. Linte, and Z.~Yaniv, ``A distance map regularized cnn for
  cardiac cine {MR} image segmentation,'' {\em Medical Physics}, vol.~46,
  no.~12, pp.~5637--5651, 2019.

\bibitem{park2019deepsdf}
J.~J. Park, P.~Florence, J.~Straub, R.~Newcombe, and S.~Lovegrove, ``Deepsdf:
  Learning continuous signed distance functions for shape representation,'' in
  {\em CVPR}, pp.~165--174, 2019.

\bibitem{hinton2015distilling}
G.~Hinton, O.~Vinyals, and J.~Dean, ``Distilling the knowledge in a neural
  network,'' {\em arXiv preprint arXiv:1503.02531}, 2015.

\bibitem{romero2014fitnets}
A.~Romero, N.~Ballas, S.~E. Kahou, A.~Chassang, C.~Gatta, and Y.~Bengio,
  ``Fitnets: Hints for thin deep nets,'' {\em arXiv preprint arXiv:1412.6550},
  2014.

\bibitem{liu2019structured}
Y.~Liu, K.~Chen, C.~Liu, Z.~Qin, Z.~Luo, and J.~Wang, ``Structured knowledge
  distillation for semantic segmentation,'' in {\em CVPR}, 2019.

\bibitem{yu2019annotation}
F.~Yu, J.~Zhao, Y.~Gong, Z.~Wang, Y.~Li, F.~Yang, B.~Dong, Q.~Li, and L.~Zhang,
  ``Annotation-free cardiac vessel segmentation via knowledge transfer from
  retinal images,'' in {\em MICCAI}, 2019.

\bibitem{li2020dual}
K.~Li, S.~Wang, L.~Yu, and P.-A. Heng, ``Dual-teacher++: Exploiting
  intra-domain and inter-domain knowledge with reliable transfer for cardiac
  segmentation,'' {\em IEEE Transactions on Medical Imaging}, 2020.

\bibitem{he2019dpa}
Y.~He, G.~Yang, Y.~Chen, Y.~Kong, J.~Wu, L.~Tang, X.~Zhu, J.-L. Dillenseger,
  P.~Shao, S.~Zhang, {\em et~al.}, ``Dpa-densebiasnet: Semi-supervised {3D}
  fine renal artery segmentation with dense biased network and deep priori
  anatomy,'' in {\em MICCAI}, pp.~139--147, Springer, 2019.

\bibitem{zhou2019models}
Z.~Zhou, V.~Sodha, M.~M.~R. Siddiquee, R.~Feng, N.~Tajbakhsh, M.~B. Gotway, and
  J.~Liang, ``Models genesis: Generic autodidactic models for {3D} medical
  image analysis,'' in {\em MICCAI}, pp.~384--393, Springer, 2019.

\bibitem{yang2020nuset}
L.~Yang, R.~P. Ghosh, J.~M. Franklin, S.~Chen, C.~You, R.~R. Narayan, M.~L.
  Melcher, and J.~T. Liphardt, ``Nuset: A deep learning tool for reliably
  separating and analyzing crowded cells,'' {\em PLoS computational biology},
  2020.

\bibitem{chen2021deep}
X.~Chen, S.~Sun, N.~Bai, K.~Han, Q.~Liu, S.~Yao, H.~Tang, C.~Zhang, Z.~Lu,
  Q.~Huang, {\em et~al.}, ``A deep learning-based auto-segmentation system for
  organs-at-risk on whole-body computed tomography images for radiation
  therapy,'' {\em Radiotherapy and Oncology}, 2021.

\bibitem{shanlin2022CVPR}
S.~Shanlin, H.~Kun, K.~Deying, T.~Hao, Y.~Xiangyi, and X.~Xiaohui,
  ``Topology-preserving shape reconstruction and registration via neural
  diffeomorphic flow,'' in {\em CVPR}, 2022.

\bibitem{tang2019clinically}
H.~Tang, X.~Chen, Y.~Liu, Z.~Lu, J.~You, M.~Yang, S.~Yao, G.~Zhao, Y.~Xu,
  T.~Chen, {\em et~al.}, ``Clinically applicable deep learning framework for
  organs at risk delineation in ct images,'' {\em Nature Machine Intelligence},
  2019.

\bibitem{tang2021spatial}
H.~Tang, X.~Liu, K.~Han, X.~Xie, X.~Chen, H.~Qian, Y.~Liu, S.~Sun, and N.~Bai,
  ``Spatial context-aware self-attention model for multi-organ segmentation,''
  in {\em WACV}, 2021.

\bibitem{zhang2018fully}
X.~Zhang, D.~G. Martin, M.~Noga, and K.~Punithakumar, ``Fully automated left
  atrial segmentation from mr image sequences using deep convolutional neural
  network and unscented kalman filter,'' in {\em International Conference on
  Bioinformatics and Biomedicine}, IEEE, 2018.

\bibitem{zhang2021fully}
X.~Zhang, M.~Noga, D.~G. Martin, and K.~Punithakumar, ``Fully automated left
  atrium segmentation from anatomical cine long-axis mri sequences using deep
  convolutional neural network with unscented kalman filter,'' {\em Medical
  Image Analysis}, 2021.

\bibitem{tang2019nodulenet}
H.~Tang, C.~Zhang, and X.~Xie, ``Nodulenet: Decoupled false positive reduction
  for pulmonary nodule detection and segmentation,'' in {\em MICCAI}, 2019.

\bibitem{zhang2020fully}
X.~Zhang, M.~Noga, and K.~Punithakumar, ``Fully automated deep learning based
  segmentation of normal, infarcted and edema regions from multiple cardiac mri
  sequences,'' in {\em Myocardial Pathology Segmentation Combining
  Multi-Sequence CMR Challenge}, pp.~82--91, Springer, 2020.

\bibitem{sun2020attentionanatomy}
S.~Sun, Y.~Liu, N.~Bai, H.~Tang, X.~Chen, Q.~Huang, Y.~Liu, and X.~Xie,
  ``Attentionanatomy: A unified framework for whole-body organs at risk
  segmentation using multiple partially annotated datasets,'' in {\em ISBI},
  IEEE, 2020.

\bibitem{yan2022after}
X.~Yan, H.~Tang, S.~Sun, H.~Ma, D.~Kong, and X.~Xie, ``After-unet: Axial fusion
  transformer unet for medical image segmentation,'' in {\em WACV}, 2022.

\bibitem{tang2021recurrent}
H.~Tang, X.~Liu, S.~Sun, X.~Yan, and X.~Xie, ``Recurrent mask refinement for
  few-shot medical image segmentation,'' in {\em ICCV}, 2021.

\bibitem{ma2020distance}
J.~Ma, Z.~Wei, Y.~Zhang, Y.~Wang, R.~Lv, C.~Zhu, C.~Gaoxiang, J.~Liu, C.~Peng,
  L.~Wang, {\em et~al.}, ``How distance transform maps boost segmentation cnns:
  an empirical study,'' in {\em Medical Imaging with Deep Learning}, PMLR,
  2020.

\bibitem{castillo2020auxiliary}
J.~Castillo-Navarro, B.~Le~Saux, A.~Boulch, and S.~Lef{\`e}vre, ``On auxiliary
  losses for semi-supervised semantic segmentation,'' in {\em ECML PKDD}, 2020.

\bibitem{murugesan2019psi}
B.~Murugesan, K.~Sarveswaran, S.~M. Shankaranarayana, K.~Ram, J.~Joseph, and
  M.~Sivaprakasam, ``Psi-net: Shape and boundary aware joint multi-task deep
  network for medical image segmentation,'' in {\em EMBC}, IEEE, 2019.

\bibitem{xia2020uncertainty}
Y.~Xia, D.~Yang, Z.~Yu, F.~Liu, J.~Cai, L.~Yu, Z.~Zhu, D.~Xu, A.~Yuille, and
  H.~Roth, ``Uncertainty-aware multi-view co-training for semi-supervised
  medical image segmentation and domain adaptation,'' {\em Medical Image
  Analysis}, 2020.

\bibitem{zheng2019semi}
H.~Zheng, L.~Lin, H.~Hu, Q.~Zhang, Q.~Chen, Y.~Iwamoto, X.~Han, Y.-W. Chen,
  R.~Tong, and J.~Wu, ``Semi-supervised segmentation of liver using adversarial
  learning with deep atlas prior,'' in {\em MICCAI}, pp.~148--156, Springer,
  2019.

\bibitem{zhuang2019self}
X.~Zhuang, Y.~Li, Y.~Hu, K.~Ma, Y.~Yang, and Y.~Zheng, ``Self-supervised
  feature learning for {3D} medical images by playing a rubik’s cube,'' in
  {\em MICCAI}, pp.~420--428, Springer, 2019.

\bibitem{taleb20203d}
A.~Taleb, W.~Loetzsch, N.~Danz, J.~Severin, T.~Gaertner, B.~Bergner, and
  C.~Lippert, ``{3D} self-supervised methods for medical imaging,'' in {\em
  NeurIPS}, pp.~18158--18172, 2020.

\bibitem{you2022class}
C.~You, R.~Zhao, F.~Liu, S.~Chinchali, U.~Topcu, L.~Staib, and J.~S. Duncan,
  ``Class-aware generative adversarial transformers for medical image
  segmentation,'' {\em arXiv preprint arXiv:2201.10737}, 2022.

\bibitem{hadsell2006dimensionality}
R.~Hadsell, S.~Chopra, and Y.~LeCun, ``Dimensionality reduction by learning an
  invariant mapping,'' in {\em CVPR}, 2006.

\bibitem{doersch2015unsupervised}
C.~Doersch, A.~Gupta, and A.~A. Efros, ``Unsupervised visual representation
  learning by context prediction,'' in {\em ICCV}, pp.~1422--1430, 2015.

\bibitem{noroozi2016unsupervised}
M.~Noroozi and P.~Favaro, ``Unsupervised learning of visual representations by
  solving jigsaw puzzles,'' in {\em ECCV}, pp.~69--84, Springer, 2016.

\bibitem{wu2018unsupervised}
Z.~Wu, Y.~Xiong, X.~Y. Stella, and D.~Lin, ``Unsupervised feature learning via
  non-parametric instance discrimination,'' in {\em CVPR}, 2018.

\bibitem{tian2019contrastive}
Y.~Tian, D.~Krishnan, and P.~Isola, ``Contrastive multiview coding,'' {\em
  arXiv preprint arXiv:1906.05849}, 2019.

\bibitem{misra2020self}
I.~Misra and L.~v.~d. Maaten, ``Self-supervised learning of pretext-invariant
  representations,'' in {\em CVPR}, pp.~6707--6717, 2020.

\bibitem{Federici2020Learning}
M.~Federici, A.~Dutta, P.~Forré, N.~Kushman, and Z.~Akata, ``Learning robust
  representations via multi-view information bottleneck,'' in {\em ICLR}, 2020.

\bibitem{chen2021self}
N.~Chen, C.~You, and Y.~Zou, ``Self-supervised dialogue learning for spoken
  conversational question answering,'' in {\em INTERSPEECH}, 2021.

\bibitem{you2021self}
C.~You, N.~Chen, and Y.~Zou, ``Self-supervised contrastive cross-modality
  representation learning for spoken question answering,'' {\em arXiv preprint
  arXiv:2109.03381}, 2021.

\bibitem{oord2018representation}
A.~v.~d. Oord, Y.~Li, and O.~Vinyals, ``Representation learning with
  contrastive predictive coding,'' {\em arXiv preprint arXiv:1807.03748}, 2018.

\bibitem{wang2015unsupervised}
X.~Wang and A.~Gupta, ``Unsupervised learning of visual representations using
  videos,'' in {\em ICCV}, pp.~2794--2802, 2015.

\bibitem{you2020data}
C.~You, N.~Chen, F.~Liu, D.~Yang, and Y.~Zou, ``Towards data distillation for
  end-to-end spoken conversational question answering,'' {\em arXiv preprint
  arXiv:2010.08923}, 2020.

\bibitem{you2020contextualized}
C.~You, N.~Chen, and Y.~Zou, ``Contextualized attention-based knowledge
  transfer for spoken conversational question answering,'' in {\em
  INTERSPEECH}, 2021.

\bibitem{you2021mrd}
C.~You, N.~Chen, and Y.~Zou, ``{MRD-N}et: {M}ulti-{M}odal {R}esidual
  {K}nowledge {D}istillation for {S}poken {Q}uestion {A}nswering,'' in {\em
  {IJCAI}}, 2021.

\bibitem{you2021knowledge}
C.~You, N.~Chen, and Y.~Zou, ``Knowledge distillation for improved accuracy in
  spoken question answering,'' in {\em ICASSP}, 2021.

\bibitem{ma2020undistillable}
H.~Ma, T.~Chen, T.-K. Hu, C.~You, X.~Xie, and Z.~Wang, ``Undistillable: Making
  a nasty teacher that cannot teach students,'' in {\em ICLR}, 2020.

\bibitem{ma2021good}
H.~Ma, T.~Chen, T.-K. Hu, C.~You, X.~Xie, and Z.~Wang, ``Good students play big
  lottery better,'' {\em arXiv preprint arXiv:2101.03255}, 2021.

\bibitem{furlanello2018born}
T.~Furlanello, Z.~Lipton, M.~Tschannen, L.~Itti, and A.~Anandkumar, ``Born
  again neural networks,'' in {\em ICML}, 2018.

\bibitem{yang2018knowledge}
C.~Yang, L.~Xie, S.~Qiao, and A.~Yuille, ``Knowledge distillation in
  generations: More tolerant teachers educate better students,'' {\em arXiv
  preprint arXiv:1805.05551}, 2018.

\bibitem{li2017learning}
Y.~Li, J.~Yang, Y.~Song, L.~Cao, J.~Luo, and L.-J. Li, ``Learning from noisy
  labels with distillation,'' in {\em ICCV}, pp.~1910--1918, 2017.

\bibitem{xie2018improving}
J.~Xie, B.~Shuai, J.-F. Hu, J.~Lin, and W.-S. Zheng, ``Improving fast
  segmentation with teacher-student learning,'' {\em arXiv preprint
  arXiv:1810.08476}, 2018.

\bibitem{zhang2018deep}
Y.~Zhang, T.~Xiang, T.~M. Hospedales, and H.~Lu, ``Deep mutual learning,'' in
  {\em CVPR}, pp.~4320--4328, 2018.

\bibitem{chen2021semi}
X.~Chen, Y.~Yuan, G.~Zeng, and J.~Wang, ``Semi-supervised semantic segmentation
  with cross pseudo supervision,'' in {\em CVPR}, 2021.

\bibitem{vu2019advent}
T.-H. Vu, H.~Jain, M.~Bucher, M.~Cord, and P.~P{\'e}rez, ``Advent: Adversarial
  entropy minimization for domain adaptation in semantic segmentation,'' in
  {\em CVPR}, pp.~2517--2526, 2019.

\bibitem{verma2019interpolation}
V.~Verma, K.~Kawaguchi, A.~Lamb, J.~Kannala, Y.~Bengio, and D.~Lopez-Paz,
  ``Interpolation consistency training for semi-supervised learning,'' in {\em
  IJCAI}, 2019.

\bibitem{klambauer2017self}
G.~Klambauer, T.~Unterthiner, A.~Mayr, and S.~Hochreiter, ``Self-normalizing
  neural networks,'' in {\em NeurIPS}, pp.~972--981, 2017.

\bibitem{roth2016data}
H.~R. Roth, A.~Farag, E.~Turkbey, L.~Lu, J.~Liu, and R.~M. Summers, ``Data from
  pancreas-ct. the cancer imaging archive,'' 2016.

\bibitem{zhou2019prior}
Y.~Zhou, Z.~Li, S.~Bai, C.~Wang, X.~Chen, M.~Han, E.~Fishman, and A.~L. Yuille,
  ``Prior-aware neural network for partially-supervised multi-organ
  segmentation,'' in {\em ICCV}, pp.~10672--10681, 2019.

\bibitem{ssl4mis2020}
X.~Luo, ``{SSL4MIS}.'' \url{https://github.com/HiLab-git/SSL4MIS}, 2020.

\end{thebibliography}

\end{document}